\documentclass[a4paper,11pt]{article}
\usepackage[pdftex]{graphicx}
\usepackage[numbers]{natbib}
\usepackage[pdftex, pdftitle={Quantum Consciousness Soccer Simulator},
pdfauthor={Norbert Batfai},
pdfkeywords={Soccer Simulation, Human Consciousness, Machine Consciousness, Soccer Consciousness},
pdfstartview=FitB]{hyperref}
\usepackage{amssymb,amsmath,amsthm,amsfonts}
\newtheorem{theorem}{Theorem}
\newtheorem{definition}{Definition}
\newtheorem{notation}{Notation}

\newcommand{\keywords}[1]{\par\addvspace\baselineskip\noindent\textbf{Keywords: }\textit{#1}.}

\author{Norbert B\'atfai
\\University of Debrecen
\\Department of Information Technology
\\\texttt{batfai.norbert@inf.unideb.hu}}
\title{Quantum Consciousness Soccer Simulator}
\begin{document}
\maketitle

\begin{abstract}
In cognitive sciences it is not uncommon to use various games effectively. For example, in artificial intelligence,  the RoboCup \cite{robocup} initiative was to set up to catalyse research on the field of autonomous agent technology. In this paper, we introduce a similar soccer simulation initiative to try to investigate a model of human consciousness and a notion of reality in the form of a cognitive problem. In addition, for example, the home pitch advantage and the objective role of the supporters could be naturally described and discussed  in terms of this new soccer simulation model.
\end{abstract}

%
% Introduction
%
\section{Introduction}

The robot soccer, or commonly called RoboCup, is a standard AI problem for catalyzing research on the field of autonomous agent technology \cite{robocup}. In RoboCup, there are several different kinds of leagues.  Currently, in the case of RoboCup 2D Soccer Simulation League (2D RCSS),  all aspects of the game of the world's best teams are quite real if compared to﻿ the matches among various humanoid teams,  while the same cannot be said of the case of the other leagues of RoboCup. 

In 2D soccer simulations, the \texttt{rcssserver} \cite{rcssserver} establishes the reality of the simulated soccer world. Through UDP/IP, client agents have connected to this simulated reality. But they are taking part in the establishment of reality only through the \texttt{rcssserver} using RCSS protocol  \cite{rcssmanual}.  Following this protocol, the client agents receive their sensory input from the \texttt{rcssserver}, then send back a "conscious" response, and this cycle takes place repeatedly in the usual manner in autonomous agent technologies.

In contrast with this, we would like to develop a new concept for simulation of soccer in that the client agents are more directly related to the establishment of reality. The new soccer simulation environment is partly inspired  
by several interpretations of quantum mechanics \cite{emperor, wigner, neumann, neumannwigner, cat, cat2, vasy}, for example Hugh Everett's Many-worlds, Wheeler's participatory universe, Many-minds, Copenhagen or Neumann and Wigner's interpretations. But it is important to note that we are only at the popular science level of understanding of these issues and  the quantum mechanical inspiration will play no part in the next chapters. However, in the case of soccer, some interpretations of quantum mechanics may enable, in theory, that all actions of all client agents might be real by representing forks in the simulation process. In this case, the known question is that how the client agents are to be selected such that they play the same match. In philosophical level, it may be supposed that the nature has already done this selection in the real world. But in the simulation, we have to make it ourselves. In order to fulfill this, drifting away from the many-worlds and many-minds interpretations and towards the Copenhagen as well as Neumann and Wigner's interpretations, we introduce a scheduler to select only one among many parallel realities. It will be called Quantum Consciousness Soccer Simulator, or briefly QCSS.

The choice of the name "Quantum Consciousness Soccer Simulator" is suggested by the Penrose-Hameroff Orch OR (Orchestrated Objective Reduction) model of consciousness \cite{orchor, orchor2, orchor3,  orchor4}. This amazing Orch OR model of consciousness is based on quantum mechanics.

In the next section, we define the terms of QCSS.  We just hope that we can specify an interesting (standard) cognitive problem, as RoboCup has become in the field of AI in the past 15 years.

%
% The Quantum Consciousness Soccer Simulator
%
\section{The Quantum Consciousness Soccer Simulator}

The new concept of playing soccer introduced in this section is entirely based on assumptions rather than on any direct observations and  experiences.

In general, six types of roles will be distinguished in the simulation environment: players, referees, coaches, managers, supporters and couch potato supporters. Actually, in this paper, we focus only on two types of roles: players and supporters. The members of all roles are autonomous software agents, for example, in the sense of the paper \cite{autonomousagents}. 
In the following, we will use the terminology "autonomous soccer agents".
Any autonomous soccer agents are characterized by a function $\operatorname{w}$,  referred to as the power of will function. 

$$\mbox{For example, } p \in R_{player}, \operatorname{w}(p)=1,  \sum_{s \in R_{supporter}}\operatorname{w}(s) \le 1.$$

This function shows how strong the influence of a role during  the establishment of reality. 
It may be interesting to note that the aforementioned $\sum\operatorname{w}(s)=1$ may be interpreted as the supporters are the 12th player.

Throughout the following, the set $R = R_{player} \cup R_{supporter}$ denotes a given final set of  members of all roles.

%
% vector of play
%
\begin{definition}[state vector of play]
Let $p_i, q_i \in R_{player}$ be autonomous soccer agents (players) for $i=1, \dots , 11$.
The 25-tuple
\begin{align}( (x_{ball}, y_{ball}), 
& (x_{p_1}, y_{p_1}), \dots
(x_{p_{11}}, y_{p_{11}}), \\
& (x_{q_1}, y_{q_1}), \dots (x_{q_{11}}, y_{q_{11}}), \nonumber
t \in \{home, guest\}, 
j \in \{1, \dots , 11\}
)
\end{align}
is called the state vector of the simulation of playing soccer, where the tuple's first component is the position of the ball and then the next components are the positions of the players $p_i$ and $q_i ,$  $i=1, \dots , 11$. Finally, the last two numbers denote the  ball-possessing team and the ball-possessing player (or more precisely, the player who touched the ball last). 
\end{definition}

This 25-tuple will describe the simulation steps. 
It is interesting to note that the FerSML (Football(er) Simulation Markup Language, introduced in \cite{fersml} and implemented in \cite{sffersml}) simulation steps could be described with a similar model of states, because it is based on tactical lineups (i.e. distinguished positions of the players) and the ball-possessing player's method of passing.

%
% receiving and sending state vectors
%
\begin{notation}[receiving and sending state vectors]
Let $r \in R$ be an autonomous soccer agent. The notation $r\gets$ denotes that the agent $r$ receives a state vector from the QCSS scheduler. The $r\gets$ is also the received state vector itself. Symmetrically, the $r\to$ denotes that the agent $r$ sends a state vector to the QCSS scheduler and it is the sent state vector, too. Finally, $r\circlearrowleft$ denotes that the agent $r$ sends a state vector to itself and it is the sent-received state vector as well. 
\end{notation}

%
% the QCSS scheduler
%
\begin{definition}[the QCSS scheduler]
Let $p_i \in R_{player}$ and $s_j \in R_{supporter}$  be autonomous soccer agents.
The QCSS scheduler is an algorithm which, from a given input $p_i\to$ and $s_j\to$ selects only one $r\gets$ state vector of play. 
\end{definition}

%
% a representation of the simulation steps
%
\begin{notation}[a representation of the simulation steps]
Let $r_l \in R$ be an autonomous soccer agent in the role of player or supporter $(l=1, \dots, n)$. The following notation shows a simulation step.
At the time $t$, all agents has received the same input state vector $r\overset{t}{\gets}$. Then they have begun their own inner simulation steps.
\[ \begin{array}{cccccccc}
\hline
\multicolumn{8}{c}{\textrm{reality: }r\gets = r_l\gets = r\overset{t}{\gets}  (l=1, \dots, n)}\\
\hline
r\gets&  r\gets& \dots & r\gets & \dots &  r\gets  & \dots &  r\gets  \\
r_1\circlearrowleft &  r_2\to & \dots & r_i\circlearrowleft  & \dots &  r_j\circlearrowleft   & \dots &  r_n\circlearrowleft   \\
r_1\circlearrowleft &  & \dots & r_i\circlearrowleft  & \dots &  r_j\circlearrowleft   & \dots &  r_n\circlearrowleft   \\
r_1\to &  & \dots & r_i\circlearrowleft  & \dots &  r_j\circlearrowleft   & \dots &  r_n\circlearrowleft   \\
 &  & \dots & r_i\circlearrowleft  & \dots &  r_j\circlearrowleft   & \dots &  r_n\to   \\
 &  & \dots & r_i\circlearrowleft  & \dots &  r_j\circlearrowleft   & \dots &   \\
 &  & \dots & r_i\circlearrowleft  & \dots &  r_j\to   & \dots &    \\
 &  & \dots & \text{\small timeout } & \dots &  & \dots &   \\
\hline
\multicolumn{8}{c}{\textrm{selecting the k-th state vector, reality: }  r^{,}\gets = r^{,}\overset{t+1}{\gets} = r_k\overset{t}{\to} }\\
\hline
r^{,}\gets&  r^{,}\gets& \dots & r^{,}\gets & \dots &  r^{,}\gets  & \dots &  r^{,}\gets  \\
 \end{array} \] 
The reality $r^{,}\gets = r^{,}\overset{t+1}{\gets}$ of the next time moment will be simply selected from the state vectors $r_l\to = r_l\overset{t}{\to},   (l=1, \dots, n)$ by the QCSS scheduler.
\end{notation}

It is important to note that the QCSS scheduler has not executed any simulation steps because this is only done by the agents. In addition, the QCSS scheduler also set the value of the function "power of will" of agents. To be more precise, the "soccer consciousness" function modifies the function of the power of will.

%
% the power of will function
%
\begin{definition}[power of will functions]
A function $w: R_{player} \cup R_{supporter} \to \mathbb{R}$ is called a power of will function if it satisfies the conditions $\sum_{p \in R_{player}}\operatorname{w}(p)=\vert R_{player}\vert$ and $\sum_{s \in R_{supporter}}\operatorname{w}(s) \le 1$.
\end{definition}

%
% the soccer consciousness function
%
\begin{definition}[soccer consciousness functions]
Now and in the following, let $S$ denote the set of the all possible state vectors. The $sc: S \times S \to  \mathbb{R}$,
$$ 
sc(r\to, r\gets) 
= \left\{ \begin{array}{l}
\frac{w(r)}{d(r\to, r\gets)} \textrm{, if } d(r\to, r\gets) \ge 0\\
max\{sc(q\to, q\gets)\vert r, q \in R_x) \}  \textrm{, if } d(r\to, r\gets) = 0
\end{array} 
\right .
$$
or more precisely, 
$$ 
sc(r\overset{t-1}{\to}, r\overset{t}{\gets}) 
= \left\{ \begin{array}{l}
\frac{w(r)}{d(r\overset{t-1}{\to}, r\overset{t}{\gets})} \textrm{, if } d(r\overset{t-1}{\to}, r\overset{t}{\gets}) \ge 0\\
max\{sc(q\overset{t-1}{\to}, q\overset{t}{\gets})\vert r, q \in R_x) \}  \textrm{, if } d(r\overset{t-1}{\to}, r\overset{t}{\gets}) = 0
\end{array} 
\right .
$$
function is referred to as a soccer consciousness function, where $d$ is the Euclidean distance. In that theoretical case, when $d(r\to, r\gets) = 0$ for all $ r \in R_x$, let $sc(r\to, r\gets)$ equal to $w(r)$, where $x$ denotes the role of the agent $r$.
\end{definition}

Here, the values of this trivial function $sc$ simply depends only on the distance between the sent and the finally selected state vectors. But in general, the purpose of the functions like $sc$ are to tell how the predicted $r\to$ of a client agent $r$ differs from the $ r\gets$ selected in the reality, in the sense of the paper \cite{consci}. That is, a good soccer consciousness function (machine consciousness function) should measure to what extent can an agent see the future. Or, in the terminology of the mentioned paper \cite{consci}, it investigates how conscious or intuitive an agent is.

%
% the selection procedure of the  QCSS scheduler
%
\begin{definition}[a selection procedure of the  QCSS scheduler]
Let $r_l \in R$ be an autonomous soccer agent in the role of player or supporter $(l=1, \dots, n)$. At the time $t+1$, 
the $r\gets$ will be selected from the probability distribution 
\begin{align}
\mathbf{P}(r\gets = r_l\to) = 
\frac{sc(r_l\to, r_l\gets)}{\sum_{i=1}^{n}{sc(r_i\to, r_i\gets)}}, (l=1, \dots, n) \nonumber
\end{align}
by the QCSS scheduler. Or to be more precise, from the probability distribution 
\begin{align}
\mathbf{P}(r\overset{t+1}{\gets} = r_l\overset{t}{\to}) = 
\frac{sc(r_l\overset{t-1}{\to}, r_l\overset{t}{\gets})}{\sum_{i=1}^{n}{sc(r_i\overset{t-1}{\to}, r_i\overset{t}{\gets})}}, (l=1, \dots, n). \label{classical}
\end{align}
\end{definition}

\begin{theorem}
$$\sum_{i=1}^{n}{\mathbf{P}(r\gets = r_l\to)} = 1.$$
\begin{proof}
It is trivial, because the Eq. \ref{classical} is based on the classical method for computing probabilities.
\end{proof}
\end{theorem}

%
% QCSS matches
%
\begin{definition}[QCSS matches]
The 6-tuple $M = (R, k\gets, w, sc, \mathbf{P})$ is called a QCSS football match, where $\vert R_{player}\vert \le 22$, $k\gets \in S$ is a starting lineup and $\mathbf{P}$ is a selection procedure of the  QCSS scheduler. 
\end{definition}

\section{The First Reference Implementations}

In the case of RoboCup there are only players and coaches. In contrast with this, football supporters must also be handled in the newly introduced simulation environment. It gives the main difficulty of the implementation because the number of supporters may be greater than 80,000. This is only partly a technical problem, because it also raises questions of principle relating to the heterogeneous composition of supporters. Regarding the technical problem, it may be a possibility to use CUDA \cite{cuda} GPU, where device threads would be corresponded to supporters. For handling heterogeneity, we may create different archetypes like attackers, midfielders and defenders among the players. 

It is may be noted that similar difficulties will arise in handling of couch potato supporters, because their number may reach hundreds of thousands. In this case, a Java EE-based \cite{javaee} solution may be investigated.

In this chapter, we will focus only on a such type of implementation in which the evolution of the fundamentals of playing soccer will be studied. 

\subsection{An Experimental Implementation of the New Concept of Soccer}

Now an asynchronous UDP server has been written in C++ using Boost.Asio \cite{boostasio} library. It is embedded in the class \texttt{QCSSStadium}. The clients are defined in the class \texttt{QCSSAgent}. 
The state vectors are abstracted by the class \texttt{StateVector}. 
This implementation can be found at SourceForge, at URL \url{https://sourceforge.net/projects/qcss/} \cite{sfqcss}, in which we use the following modified definition of the selection procedure in the method \texttt{void QCSSStadium::select\_reality (void)}.
%
% a modified selection procedure of the  QCSS scheduler
%
\begin{definition}[a modified selection procedure of the QCSS scheduler]
Let $r_l \in R$ be an autonomous soccer agent in the role of player or supporter $(l=1, \dots, n)$. 
Let $\{{r_{j_1}}\overset{t}{\to}, \dots , {r_{j_m}}\overset{t}{\to}\}$, $m \le n$ be the set of  state vectors received to the QCSS scheduler before time $t+1$.
At the time $t+1$, 
the $r\gets$ will be selected from the probability distribution 
\begin{align}
\mathbf{P}(r\overset{t+1}{\gets} = r_l\overset{t}{\to}) = 
\frac{sc(r_l\overset{t-1}{\to}, r_l\overset{t}{\gets})}{\sum_{i=1}^{m}{sc({r_{j_i}}\overset{t-1}{\to}, {r_{j_i}}\overset{t}{\gets})}}, (l=j_1, \dots, j_m). \label{late}
\end{align}
\end{definition}

This means that agents who are late are not allowed to taking part in the selection process described by Eq. \ref{late}. If $r_l\overset{t-1}{\to} \notin \{{r_{j_1}}\overset{t-1}{\to}, \dots , {r_{j_z}}\overset{t-1}{\to}\}$ 
%then let $sc(r_l\overset{t-1}{\to}, r_l\overset{t}{\gets})$ equals to $0$.
then let $\mathbf{P}(r\overset{t+1}{\gets} = r_l\overset{t}{\to})$ equal to $0$.

Finally, we remark that the function $w$ may be also changed in time in this implementation.

\subsubsection{Further Work}

During the implementation, the introduction of some new roles, such as the ball or the pitch may be arisen, where the members of these new roles could know, for example, the Newton's equations of motion. But it would be a mistake, because, for example, the laws of the motion will be come into being by itself.

At this moment, the agents contained in the experimental implementation cannot play football. This implementation may be used only for testing performance and timing of the architecture. The next step will be to program player and supporter agents to play football. For example, the simplified algorithms of FerSML platform may be used for the (subjective) implementation of the motion of players and their passes. With minimal adaptation, the FerSML platform may be applied also to visualize the stream of the selected state vectors as a soccer match.

\section{Conclusion}

It is undoubted that this paper has focused directly on soccer, but fundamentally it suggests a lot more than simply soccer. This is an initiative to create a community of programmers who would like to assist in the development of successful QCSS-based football teams and  QCSS-based football supporter groups. We hope and believe that our new simulation concept may provide an exciting framework for studying concrete models of the establishment of reality and it may become a standard cognitive problem, like RoboCup has become in the field of AI in the past 15 years.

However, to go back to the soccer, the objective role of the supporters becomes evident in the proposed new simulation model, and this objective role might explain the home pitch advantage, because in the case of a home match, it means that many home supporters can watch the match in the stadium of the home team. So, the direct reason of home pitch advantage is simply the impact of the objective role of the home supporters.

\section{Acknowledgements}

The author would like to thank to J\'anos Komzsik, P\'eter Jeszenszky and Andr\'as Mameny\'ak for reading of the manuscript and for fixing grammatical mistakes and misspellings.

\bibliography{qcss}

\end{document}